%% file: main_IEEE.tex
\documentclass[conference]{IEEEtran}
\IEEEoverridecommandlockouts

\usepackage{cite}
\usepackage{amsmath,amssymb,amsfonts}
\usepackage{algorithmic}
\usepackage{graphicx}
\usepackage{textcomp}
\usepackage{xcolor}
\input{preamble}

\def\BibTeX{{\rm B\kern-.05em{\sc i\kern-.025em b}\kern-.08em
    T\kern-.1667em\lower.7ex\hbox{E}\kern-.125emX}}
    
\begin{document}

\title{Personalizing Retrieval using Joint Embeddings; or ``the Return of Fluffy"\\
}

\author{\IEEEauthorblockN{Bruno Korbar}
\IEEEauthorblockA{\textit{Visual Geometry Group} \\
\textit{University of Oxford}\\
Oxford, UK \\
korbar@robots.ox.ac.uk}
\and
\IEEEauthorblockN{Andrew Zisserman}
\IEEEauthorblockA{\textit{Visual Geometry Group} \\
\textit{University of Oxford}\\
Oxford, UK \\
az@robots.ox.ac.uk}
}

\maketitle

\input{sec/0_abstract}


\input{sec/1_intro}
\input{sec/2_related}
\input{sec/3_model}
\input{sec/4_data}
\input{sec/5_results}

{
    \small
    \bibliographystyle{IEEEtranS}
    \bibliography{main, vgg_local}
}
\newpage

\appendix

\input{supp_sec/1_localfeatures}

\input{supp_sec/2_results}
\input{supp_sec/4_datasets}

\end{document}

%% file: preamble.tex
\usepackage{graphicx}
\usepackage{booktabs}
\usepackage{multirow}
\usepackage{bbm}
\usepackage{subcaption}
\usepackage{amssymb}
\usepackage{wrapfig}
\usepackage{breqn}
\usepackage[inline]{enumitem}

\usepackage{xspace}

\makeatletter
\DeclareRobustCommand\onedot{\futurelet\@let@token\@onedot}
\def\@onedot{\ifx\@let@token.\else.\null\fi\xspace}

\def\etal{\emph{et al}\onedot}
\makeatother

%% file: sec/0_abstract.tex
\begin{abstract}
The goal of this paper is to be able to retrieve images using a
compound query that combines object instance information from an
image, with a natural text description of what that object is doing or
where it is.  For example, to retrieve an image of `Fluffy the
unicorn (specified by an image) on someone's head'.  To achieve this
we design a mapping network that can `translate' from a local image
embedding (of the object instance) to a text token, such that the
combination of the token and a natural language query is suitable for
CLIP style text encoding, and image retrieval.
Generating a text token in this manner involves a simple training procedure, that only needs
to be performed once for each object instance. 
We show that our approach of using a trainable mapping network, termed $\pi$-map, together with {\em frozen} CLIP text and image encoders, improves the state of the art on two benchmarks designed to assess personalized retrieval.
\end{abstract}

%% file: sec/1_intro.tex
\section{Introduction}
\label{sec:intro}

Large-scale pre-trained vision-language models (VLMs) alleviated the need for training task-specific models due to their emerging capability for both intra- and cross-modal retrieval. By enforcing the alignment of text and images, these models allow us to classify objects and scenes, retrieve relevant images given a textual description, and even spatially locate specific objects in an image. However, in practical uses, we are often interested in searching for a specific ``thing'' in an image. On our phones, we may have hundreds of images of dogs, but we may only be interested in one specific dog -- our dog ``Chia''. Searching our library for ``My dog Chia with a stick", since the VLMs have no knowledge of our dog Chia,  might return either a generic dog or, for example, chia seeds. But what if we want to `teach' a VLM what ``my dog Chia'' refers to? Given the name of the dog and a few template images, can we `teach' a VLM to recognise our dog? 

\begin{figure}
    \centering
    \includegraphics[width=0.45\textwidth]{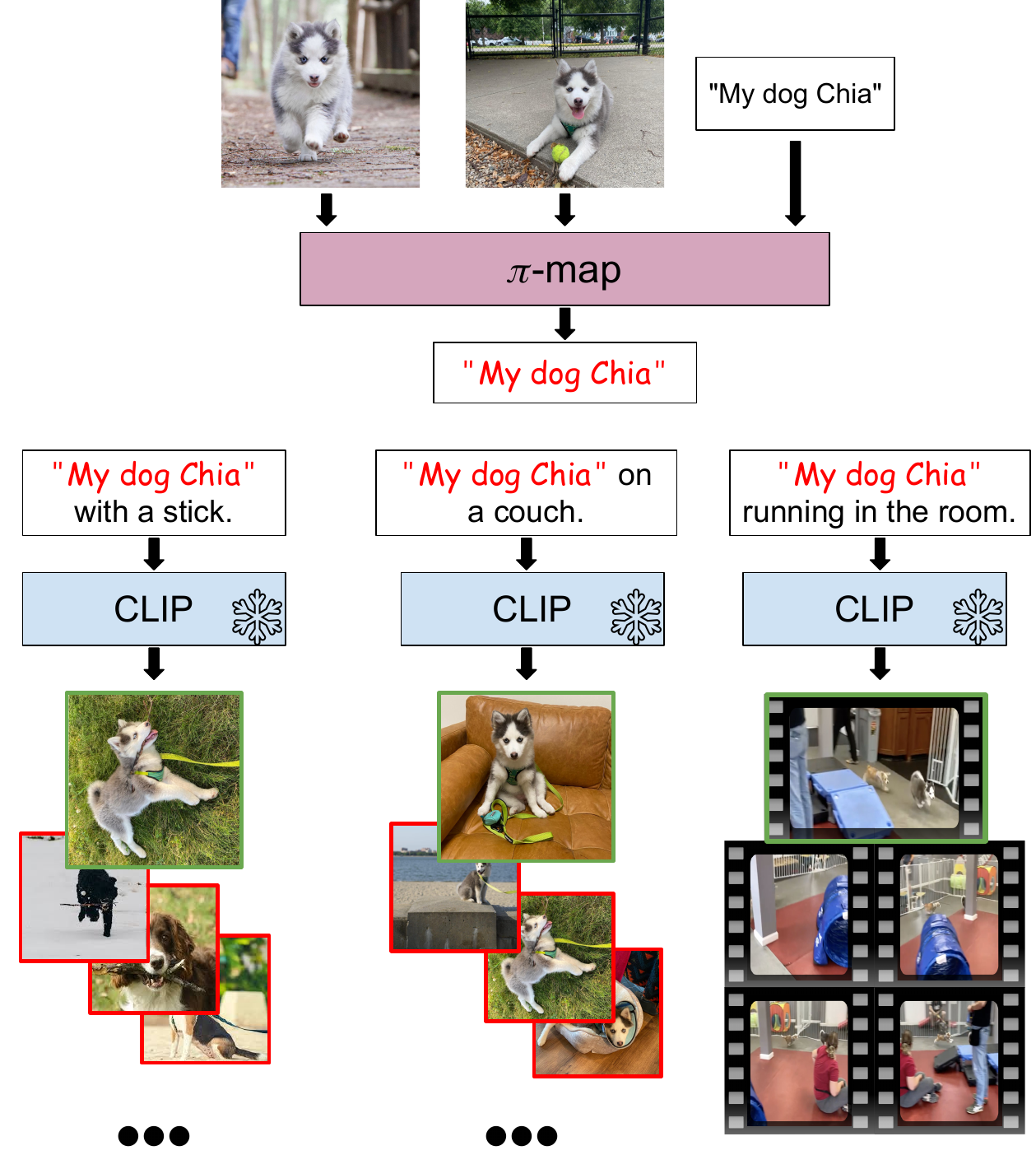}
    \caption{\small Given a few example images of an instance, our $\pi$-map model learns a personalised text embedding for the instance ('``My dog Chia''). This text embedding can then be composed with free form text queries to search amongst a dataset of images or within video frames.}
    \label{fig:teaser}
    \vspace{-10pt}
\end{figure}

%
%
In prior work, this problem has been referred to as the ``personalization'' of VLMs~\cite{eccv2022_palavra_cohen, yeh2023meta}. 

The great advantage of achieving this personalization is that we then can deploy the compositional power of the VLM, and search for ``my dog Chia" carrying out various activities and in different environments simply by writing our query as a natural language sentence, as illustrated in Fig~\ref{fig:teaser}.
Our approach is inspired by the language model's almost infinite expressively; given a specific-enough query, the large language model used in most popular VLMs \textit{should} be able to synthesise information necessary for better text-to-image retrieval.
Therefore, one could argue that the task of personalization might be expressed as learning that `my dog Chia' corresponds to `an adorable\footnote{not strictly relevant} 4-month old blue-eyed husky mix with grey inverse mask and white socks and features, about 10 inches high'.

%
%
Our approach trains a `translation' network that can map from a few example template images of the object of interest to a suitable text embedding. The text embedding is then used in query sentences for the personalized search for that object. 
We are not the first to attempt this (for example, personalization is the goal of the paper “this
is my unicorn, fluffy”~\cite{eccv2022_palavra_cohen}), and our solution builds on those of others, but our method has fewer requirements than prior work, and demonstrates superior retrieval performance. 

In terms of requirements: we are able to use {\em frozen} CLIP image and text encoders (whereas previous work fine-tuned the text encoder~\cite{Korbar22}); and by using a {\em local image embedding} we require fewer and less diverse training images than prior work for the personalization -- avoiding the failing of learning the context of the image background rather than the foreground object of interest~\cite{eccv2022_palavra_cohen}.  Furthermore, we leverage a LLM's expressivity to automatically generate caption augmentations in the language domain. Also, unlike previous work~\cite{yeh2023meta}, training does not require retrieval from a large dataset, so it is efficient.

In terms of performance, we demonstrate superior retrieval performance compared to previous methods over \textbf{two} standard benchmark datasets: this-is-my'~\cite{yeh2023meta} and  `DeepFashion2'~\cite{DeepFashion2, eccv2022_palavra_cohen}.


%% file: sec/2_related.tex
\section{Related Work}
\label{sec:related}

\paragraph{Methods for translating between image and text embeddings.}
Translation between the modalities of VLMs is a well explored topic related to the task of personalization. Mokady~\etal~\cite{mokady2021clipcap} show that a single mapping network can translate encoding from images to the text model. They fully finetune the text encoder. Alayrac~\etal propose training adapter models that map a visual input to the LLM domain using a model dubbed `Perceiver Resampler'. With such mappings, they only train adapter layers within a LLM~\cite{alayrac2022flamingo}. Li~\etal~\cite{li2023blip2} devise an even more efficient model (`Q-former') and a two-stage training method that translates any arbitrary large vision transformer into the domain of LLMs with no need for additional adapter layers. These methods have became a de-facto choice for tasks such as retrieval~\cite{alayrac2022flamingo, gorti2022x, dzabraev2021mdmmt}, and for visual question answering~\cite{li2022composing, yang2022zero, alayrac2022flamingo}.

An inherent discrepancy between the text and image embeddings has also been a subject of extensive study. Nukrai~\etal show that noise injection during the CLIP training process helps alleviate the `modality gap'~\cite{nukrai2022text}. Schrodi~\etal show that this modality gap can be attributed to as little as two dimensions within each embedding~\cite{schrodi2024two}. 

\paragraph{Test-time adaptation.}

The task of test-time adaptation (TTA) and various fine-tuning approaches are closely related to the personalization of VLMs. The goal of TTA is to leverages the unlabeled data that arrives at test time by adapting either the forward pass
or parameters of the model according to some proxy task~\cite{saenko2010adapting, alfarraevaluation}. While in the task of personalization we aim to preserve model's capabilities and only specialise it for one or two instances, the task of TTA generally requires a distribution shift of an entire model. 
Zhao~\etal~\cite{zhao2024testtime} show that VLMs can be adapted to out-of-distribution samples using reinforcement learning from CLIP's feedback. Gao~\etal~\cite{gao2024clip} show that a feature adapter can replace the need for fine-tuning VLMs. 
Wortsman~\etal~\cite{wortsman2022robust} present a robust method of fine-tuning VLMs to adapt to the test time data.

\paragraph{Personalization methods for Joint Embedding Retrieval.} 
Korbar and Zisserman~\cite{Korbar22} have explored how VLMs textual encoder can be augmented to associate a given face embedding with the corresponding name and use either interchangeably to retrieve relevant videos. This method relies on having strong face embeddings and is, therefore, limited to the domain of faces. 
Wang~\etal~\cite{wang2022dualprompt} demonstrated that expert embeddings from~\cite{Korbar22} can be replaced by a method that finds the closest generic prompt embedding to the novel class. They learn an `expert' prompt which is a function of the generic prompt. They focus on novel class discovery rather than on learning instance-specific attributes. 
Cohen~\etal~\cite{eccv2022_palavra_cohen} proposed extending VLM's language encoder's vocabulary with a newly learned token which represents a specific instance. Their method assumes a clean, manually annotated dataset of specific instances, which are seldom available. 
Yeh~\etal~\cite{yeh2023meta} learn a database of common traits of a given category (a process they dub ``meta-personalization'') and then learn a specific personalised embedding as a weighted combination of global category features. While this approach does not need a large example database (as general-category objects can be discovered automatically), it is limited to the number of common category traits it can store.

\paragraph{Compound retrieval.} 
 In compound text-to-image retrieval -- retrieval over multiple semantic axes, the focus is on specificity over each axis. Ventura~\etal developed a large-scale compound retrieval benchmark collected automatically by mining web-video captions~\cite{ventura2024covr}. Zhong~\etal~\cite{Zhong16} present a compound retrieval image dataset containing the axis `person' and `scene', while Korbar and Zisserman~\cite{Korbar22} present a video benchmark containing the axis of `person', `action', and `scene'.

 \paragraph{Conditional retrieval.}
 Similarly, a task of conditional retrieval seeks to retrieve a particular version of an image given constrained parameters, e.g.\ given a daytime image of Eiffel tower and a text prompt ``at night'', the task is to retrieve a nighttime image. Khartik~\etal~\cite{karthik2024visionbylanguage} demonstrate a method of prompt adaptation using large language models. Gu~\etal~\cite{gu2024compodiff} show superior performance on the same task using latent representation of diffusion models. While this task resembles personalisation in that it aims to retrieve a particular verison of an instance, there is no requirement to name that instance.

%% file: sec/3_model.tex
\section{Method}
\label{sec:method}

\begin{figure*}
    \centering
    \begin{subfigure}[t]{0.7\textwidth}
        \centering
        \includegraphics[width=0.95\textwidth]{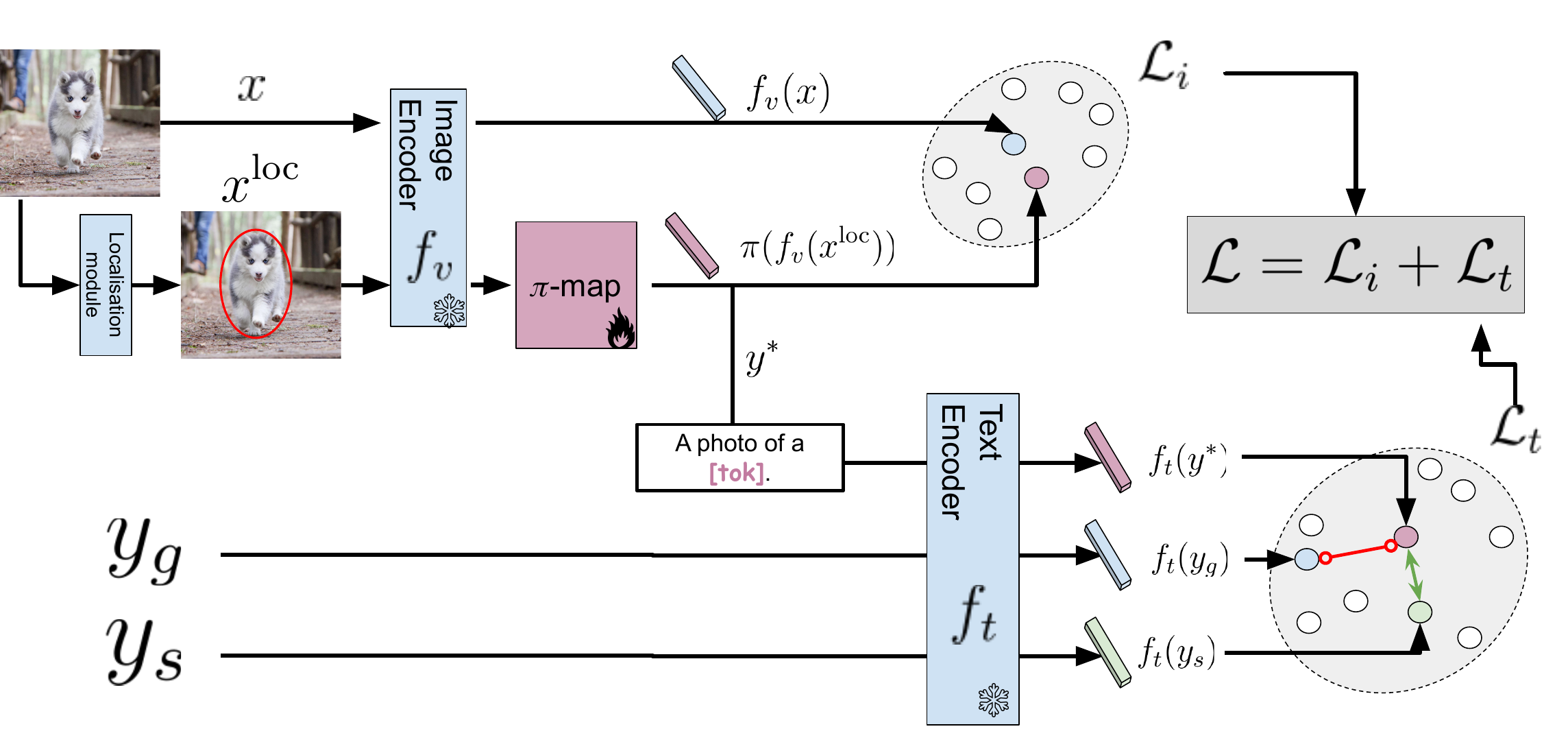}
        \caption{}
        \label{fig:training}
    \end{subfigure}
    \begin{subfigure}[t]{0.28\textwidth}
        \centering
        \includegraphics[width=\textwidth]{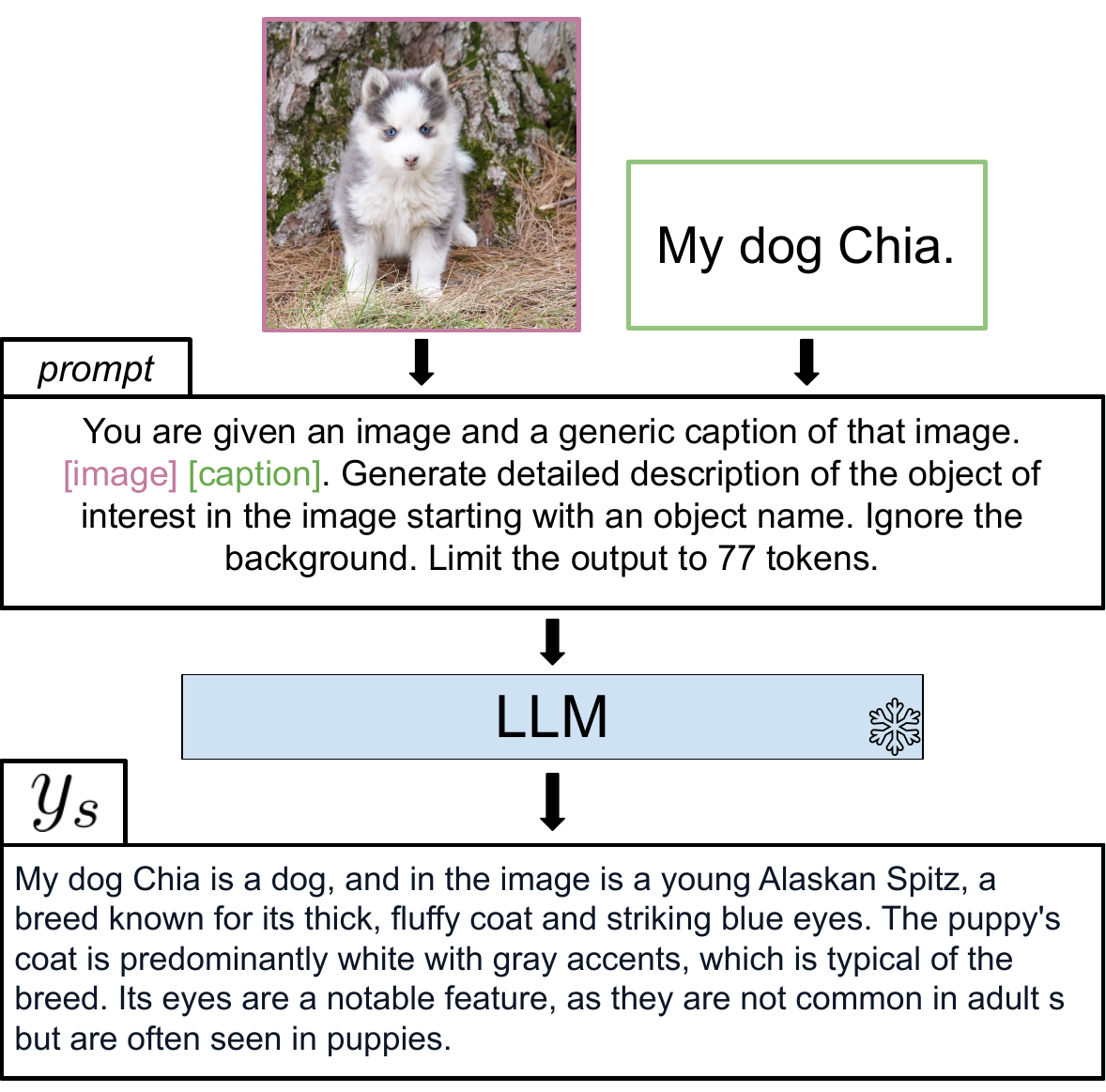}
        \caption{}
        \label{fig:textaugment}
    \end{subfigure}
    \caption{(a) {\bf Generating a text token, $y^*$, for a specific object instance}. The token $y^*$ is obtained by fine-tuning the $\pi$-map given an image $x$ of the instance and a specific text description $y_s$. The $\pi$-map is fine-tuned such that the text embedding of $y^*$ is close to the text embedding of the specific description  $y_s$ but away from the text embedding of the the generic class description $y_g$. Also, as a regularization, $y^*$ is  close to the original image embedding. The total loss is a linear combination of text embedding loss, $\mathcal{L}_t$, and the image embedding loss, $\mathcal{L}_i$. (b) Caption augmentation using an LLM~\cite{rekateam2024rekacoreflashedge} }
    \label{fig:model}
\end{figure*}

This section introduces our personalized retrieval method, describes our proposed \textbf{P}ersonalised \textbf{I}mage \textbf{E}mbedding Mapping model (PIE-map or $\pi$-map), explains the training procedure, and compares our approach to related work. 

\noindent\textbf{Overview:} Given a few example images of a specific
object, our method generates a personalized embedding (a unique text
token) through a brief one-time training process using $\pi$-map (top
part of Fig.~\ref{fig:teaser}). With this personalized embedding
(e.g., ``My dog Chia''), queries such as ``My dog Chia playing in the
park'' are formed by combining the personalized token with additional
descriptive text tokens. The CLIP text encoder processes this combined
query into an embedding, which then retrieves relevant images from a
dataset by ranking them according to their embedding similarity. An
illustration of this can be seen at the bottom of
Fig.~\ref{fig:teaser}.

\noindent\textbf{Notation: } Let $x$ be an example (template) image of the object instance, $y^*$ is the 
personalized text token we seek denoting a specific instance (`my dog Chia'), $y_g$ is text
denoting a generic object category (`dog'), and $y_s$ is text describing the specific instance. Let $f_t(\dot)$ and $f_v(\dot)$ be text
and image encoders of a CLIP VLM
respectively~\cite{radford2021learning}. The mapping  is denoted as
$\pi$. 

\noindent\textbf{Fine-tuning $\pi$-map to obtain the text token $y^*$: }
The $\pi$-map transforms visual embeddings into
representations that align closely with text embeddings; essentially
converting visual data into text-like ``words". 
To aim is to obtain a text token that represents the specific object instance (of
Chia) from the example image, but  is `distinct'
from the embedding of the general category (dogs). 

The template image  $x$ is used to obtain (i) a localized image embedding $f_v(x^{loc})$, and (ii) a 
detailed text description $y_s$ (as described below). Personalised text token $y^*$ is then obtained by minimising the
following objective function for :
$$\mathcal{L} = (1-\alpha)\mathcal{L}_t + \alpha \mathcal{L}_i$$
where $\mathcal{L}_t$ is a contrastive loss in the text embedding space that ensures that text
embedding of `an image of $y^*$' is close to the text embedding of the detailed description $y_s$, but far from
the text embedding of the  generic object category (`dog'), $y_g$.
$ \mathcal{L}_i $ is a contrastive regularization loss in the image embedding space that ensures that
the text embedding of $ y^{*} $ is close to the image embedding of the original image $x$. 
$\alpha$ is a loss balancing hyperparameter (determined by line search to be $\alpha=0.25$).

These losses are illustrated in Fig.~\ref{fig:model}, and described in detail in the following subsection.

\noindent\textbf{Inference:} 
Once the token $y^*$ has been obtained by
fine-tuning $\pi$-map (and this only needs to be done once for each
instance), then it is appended to the rest of the text of a query 
and passed through the text encoder. We measure the similarity between
the query embedding, $f_t$, and the dataset of visual features (the image embedding of each image, $f_v$) and rank based on this scalar product. An overview can be seen in Fig~\ref{fig:teaser}.

\subsection{Model details}

\noindent\textbf{Architecture: }
We use a three-layer Multi-Layer Perceptron (MLP) featuring residual
connections. Additionally, we introduce two learnable conditioning
embedding vectors that influence the outputs of the first two MLP
layers.

These conditioning vectors guide the image-to-text embedding
transformation, encouraging the embeddings to emphasize dimensions
where the most significant discrepancies between image and text
embeddings occur. As noted by~\cite{schrodi2024two}, image and text
embeddings are primarily distinguished by two key dimensions. Removing
these one or two principal components would render the embeddings
indistinguishable between modalities.

Thus, our model amplifies these critical dimensions similarly to the
way attention mechanisms operate~\cite{vaswani2017attention}. Specifically, the outputs from each of
the first two MLP layers are multiplied by conditioning vectors,
strategically enhancing the values of the embedding's most relevant
components before undergoing a final linear projection. This ensures
the model precisely focuses on dimensions essential for
differentiating between visual and textual embeddings.

\noindent\textbf{Obtaining the localised embedding.} 
By definition, $f_v(x)$ is a global embedding. Therefore, it is
sensitive to the image background and context. Say all photos of `my
dog Chia' come from a forest. The model would then be biased to all
images of a dog in a forest and might completely miss `Chia' in the
street. We
alleviate this issue by using $f_v(x^{loc})$ -- a localised
version of the embedding.

To localise the embeddings, we build on a technique by
Shtedritski~\etal~\cite{Shtedritski23} who demonstrated that drawing
a red ellipse around the area of interest focuses the  semantic
image embedding to the region within it, and Sun~\etal~\cite{clip_as_rnn}
who showed that such visually augmented image can be used as a localised
embedding for downstream tasks.

Inspired by this work, we obtain an image with a red `circle' (an
ellipse) by using a pre-trained language-guided detector to detect
objects in the image~\cite{liu2023grounding}. We empirically
demonstrate that adding a red circle around the instance to the
template images during training increases performance and
reduces the number of template images we need to form a personalised
embedding.

%
%

\noindent\textbf{Obtaining a detailed text description: }  We augment
the caption $y_s$ automatically by passing an image $x^{loc}$ and a
prompt (see illustration in Fig.~\ref{fig:textaugment}) to a large
language model~\cite{rekateam2024rekacoreflashedge} to form a detailed
text description. Previous work by Schrodi~\etal~\cite{schrodi2024two} has shown that a more expressive caption can diminish the information imbalance between text and image embeddings

\noindent\textbf{Image loss $ \mathcal{L}_i$: }
The local image embedding is mapped to
the text input using $\pi$-map to obtain $\pi(f_v(x^{loc}))$ which
becomes the basis of our personalized token $y^*$. While most of the
training is done in the text domain, we do not want
$\pi(f_v(x^{loc}))$ to collapse to an encoding of a word
`dog'. Therefore we keep a regularization loss $\mathcal{L}_i$ which
keeps $\pi(f_v(x^{loc}))$ and $f_v(x)$ close. Formally we use
contrastive loss formulation: %
$$\mathcal{L}_i(\pi(f_v(x^{loc}))) =
-log(\frac{d(\pi(f_v(x^{loc})), f_v(x))}{\sum_{n \neq x \in
\mathbbm{B}}d(\pi(f_v(x^{loc})), f_v(n)})$$ %
where $\mathbbm{B}$ is a
randomly sampled training minibatch, a distance metric is given by
$d(a,b) = exp(\frac{a^Tb / {\tau}}{||a^T||_2||b||_2 / {\tau}})$, and
$\tau = 0.07$ is a temperature hyperparameter.  Intuitively, the
regularisation loss $\mathcal{L}_i$ ensures that the projected
embedding $\pi(f_v(x^{loc}))$ does not drift from the original
embedding and thus retains its semantic information.

\noindent\textbf{Text loss $ \mathcal{L}_t$: }
This is based on three text prompts: a generic text prompt (``A photo
of a dog'', $y_g$), a specific detailed text prompt 
(``My dog chia is an alaskan
spitz...'', $y_s$), and the learned $\pi$-map embedding $y^*$. These text embeddings $y_g$, $y_s$, and
$y^* = \pi(f_v(x^{loc}))$ are then passed through CLIP text encoder
$f_t$. Since we are only training a single embedding, we want to
specialise it by making it close to $y_s$ (learning the semantic
correspondence to the detailed information), while making it less
sensitive to the more general class $y_g$ (therefore forcing our model
to extract more specialised information).  We achieve this by
optimising a contrastive objective $\mathcal{L}_t$ which ensures that
$y^*=f_t(\pi(f_v(x^{loc}))$ is similar in the embedding space to
$f_t(y_s)$ while being away from $f_t(y_g)$.
\begin{dmath} \mathcal{L}_t(\pi(f_v(x^{loc})), y_s) =
-log \frac{d(f_t(y^*),f_t(y_s))}{\sum_{y_i
\in\mathbbm{N}} d( f_t(y^*), f_t(y_i)) } \end{dmath}
where $\mathbbm{N}$ is a set of negative examples comprising all other specific and generic captions in $\mathbbm{B}$ not equal to $y_s$.

\subsection{Discussion: relation to previous methods}

Compared to the CLIP-PAD approach of~\cite{Korbar22}, we do not train the language encoder but instead use frozen versions of CLIP's image and language encoders, training only a separate module for the image-to-text translation mapping $\pi$-map. This is a great advantage as $\pi$-map can simply be `plugged in' to existing deployments of CLIP for retrieval.

Both PALVARA method of~\cite{eccv2022_palavra_cohen} and personalization approach of~\cite{yeh2023meta} learn direct text-replacement from images; ~\cite{eccv2022_palavra_cohen} from set encoding, and~\cite{yeh2023meta} by learning a linear combination of known features. This means that \textit{(a)} during the personalization stage, both of these are limited to learning from concepts they already `know'. \cite{yeh2023meta} can only represent instances that can be expressed by the linear combination of their meta categories and \cite{eccv2022_palavra_cohen} only personalises tokens learned from object detection datasets. $\pi$-map can on another hand be trained on top of an arbitrary CLIP model directly as its pre-training stage is general. 
It also means that \textit{(b)}, querying using an image token directly is impossible as \cite{yeh2023meta} requires a text prompt and \cite{eccv2022_palavra_cohen}'s prompts are fixed after the personalization stage. Because we learn direct mapping from image to text, any image can be used as a query by mapping it into the text embedding space.

\subsection{Implementation details}

\noindent\textbf{VLM details: }
For fair comparison with prior work (\cite{yeh2023meta, eccv2022_palavra_cohen}), we use OpenAI's CLIP (VIT-B/16)~\cite{radford2021learning}.

\noindent\textbf{Pre-training $\pi$-map: }
In order to initialize model and bring the modalities closer together, we pre-train $\pi$-map on ImageNet by minimizing symmetric cross-entropy loss (following~\cite{radford2021learning}) between $f_v(x)$ and $f_t(\pi(f_v(x))$. Intuitively, we are trying to `teach' $\pi$-map to map an image of a `dog', to that of the text encoding of `dog'. This is done only once.

\noindent\textbf{Initialise $\pi$-map's conditioning vectors: }We found that initialisation of conditioning vectors matters. To initialise the vectors, we first compute the image embeddings of template images and corresponding text caption, compute the absolute pairwise difference between embedding dimensions, and finally find the two dimensions with the maximum abs.\ difference. We use this difference vector, zeroing out the largest and second largest dimension respectively, before taking the softmax to initialise the first and second vectors respectively. The illustration of our model can be seen in Fig~\ref{fig:model} on the right. The effect of this initialisation scheme can be seen in Table~\ref{tbl:ablations}.

\noindent\textbf{Localisation details: }
To obtain localised image $x^{loc}$, we pass the original image and its general category to a pre-trained language-guided detection model GroundingDINO (`GroundingDINO-B')~\cite{liu2023grounding}. 
For a text prompt, GroundingDINO returns coordinate of the bounding boxes of the object. We superimpose an ellipse onto the image that passes through a centre of each side of the bounding box. 

\noindent\textbf{Obtaining a detailed text description: }
In order to augment the caption, we forward the prompt defined in Fig~\ref{fig:textaugment} and feed it to the REKA-Core model~\cite{rekateam2024rekacoreflashedge}.

\noindent\textbf{Training details: }
We pre-train the model using a batch size of 256 and a learning rate of $3e-4$ for 10 epochs. For personalization, we train the model for 50 epochs on `this-is-my' dataset, and for 80 epochs on `DeepFashion2' with a cosine annealing learning rate starting at $1e-4$ with 200 steps of linear warmup. All training is done with AdamW optimiser. In practice, a training run for learning 15 personalised tokens on `this-is-my' training set takes about 54 minutes, or $3.6$ minutes per-token on a single A4000 chip.  

\noindent {\it Using multiple example images.} 
When multiple example images are present, we randomly sample one to generate detailed description $y_s$, and keep it fixed for training across all template images. To generate the final embedding $y^*$, we average the $\pi$-map projection of them.

\noindent {\it Extension to videos. }
Keen-eyed readers would have noticed that $x$ has thus far been described as an image, but one of the datasets contains video training data. For such cases, we sub-sample a training query video to 10 frames uniformly, localise the specific instance if applicable, and encode the frames using visual encoder $f_v$. We then average embeddings to get $f_v(x)$ and $f_v(x^{loc})$.


%% file: sec/4_data.tex
\section{Datasets and Evaluation Measures}
\label{sec:datasets}

In this section, we describe the datasets used for evaluating our personalization method, as well as the evaluation measures used for each of them.

\subsection{This-is-my}
\label{subsec:thisismy}
Yeh~\etal proposed `this-is-my`~\cite{yeh2023meta} for personalised text-to-\textit{video} retrieval. The dataset consists of 104 training segments, 683 evaluation segments, and 30 test segments annotated with ten general categories (e.g.\ `dog') and 15 specific categories 
(e.g.\ `my dog Biscuit'). We use it for method development and downstream performance evaluation. \\

\noindent \textbf{Evaluation procedure: }
In order to evaluate our model, we extract CLIP image features from 30 test segments, uniformly sampling frames at 1fps following the prescribed protocol in~\cite{radford2021learning, openclip}. We embed the textual query using CLIP, 
with its tokenizer trained to recognise each of the 30 instances. We define similarity for a given video as the maximum dot product between the query and all video features. For the \textit{generic} setting (`An image of $y^*$') and report mean average precision (mAP) and mean reciprocal rank (MRR) following~\cite{yeh2023meta}. For the \textit{context}ualised setting (`A photo of $y^*$ in the car park'), there is only one correct match, hence we report MRR and recall-at-5 (R@5).  For a fair comparison with SOTA, once the training hyperparameters are set, we train the model on both the train and eval set as~\cite{yeh2023meta} train their model on both (eval set is referred to as `personalisation' dataset in their work). The embedding is formed by using 5 randomly sampled and localised frames from an eval video.

\subsection{DeepFashion2}
\label{subsec:deepfashion}
Cohen~\etal~\cite{eccv2022_palavra_cohen} proposed a modified version of DeepFashion2~\cite{DeepFashion2} for personalization purposes. 
They curated a dataset of 653 training and 221 evaluation images that have assigned one of 50 [CONCEPT] tags: e.g.\ `a white skirt', `a short dress', etc.
For the evaluation images, they collect in-context captions such as `The [CONCEPT] is facing a glass store display' (short caption) or `White cabinets, some with open drawers, are alongside and behind the [CONCEPT]' (long caption). Overall, 50 total concepts are contained in the dataset. \\

\noindent \textbf{Evaluation procedure: }
As DeepFashion2 is an image dataset, we simply encode each image with a CLIP visual model, and follow the same evaluation protocol as for the `this-is-my' dataset otherwise. We follow the benchmark setting from~\cite{yeh2023meta} and use five images to form the embedding. 

%% file: sec/5_results.tex
\section{Results}
\label{sec:results}

\begin{figure*}
    \centering
    \begin{subfigure}[l]{0.49\textwidth}
    \includegraphics[width=\textwidth]{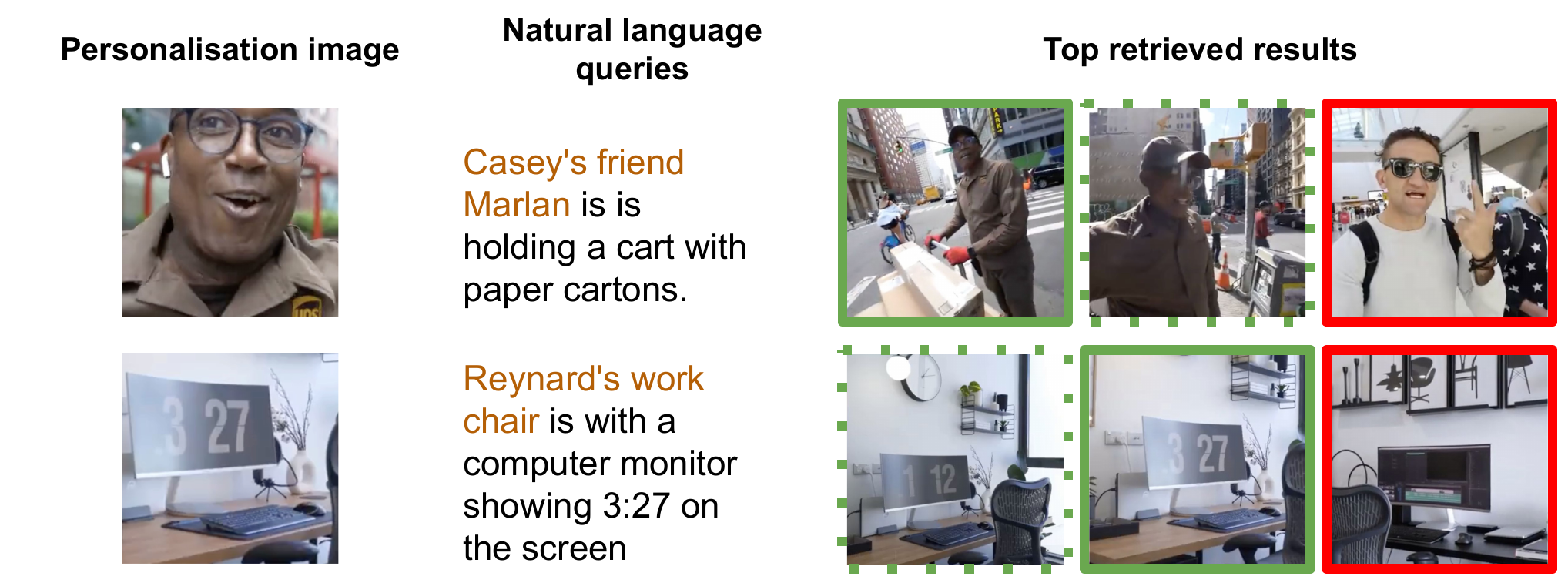}
    \end{subfigure}
    \begin{subfigure}[r]{0.49\textwidth}
        \includegraphics[width=\textwidth]{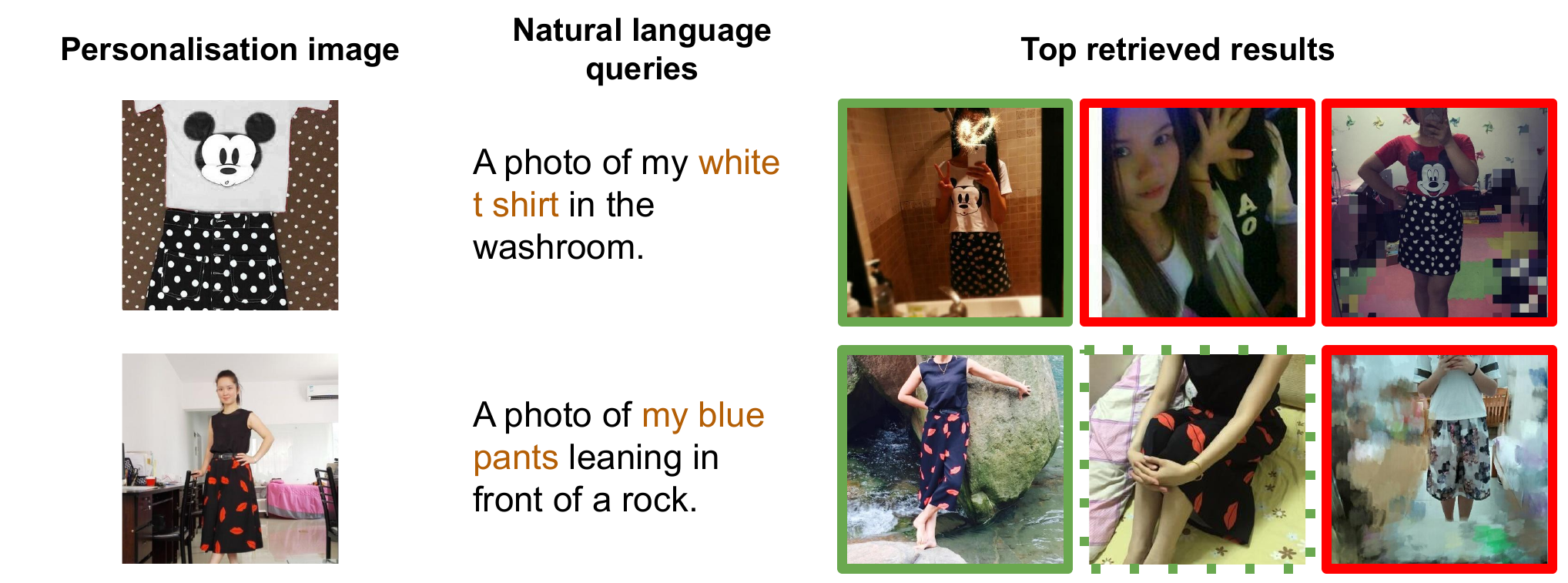}
    \end{subfigure}
    \caption{A qualitative sample of contextual retrieval sorted from left to right from this-is-my~\cite{yeh2023meta} and DeepFashion2~\cite{DeepFashion2} datasets. Green and red rectangles correspond to the correctly and incorrectly retrieved segments/images. Dotted green line shows correctly retrieved instances but in wrong setting.}
    \label{fig:retrieval}
\end{figure*}

In this section, we present the results of our method. Sec~\ref{subsec:ablations} presents various ablation studies taken into account while designing the model. We then compare our trained model with state-of-the-art (SOTA) on the personalisation datasets described in Sec~\ref{sec:datasets}: `this-is-my'~\cite{yeh2023meta} (in Sec.~\ref{subsec:thisismyRes}), and `DeepFashion2'~\cite{DeepFashion2} (in Sec~\ref{subsec:deepfashionRes}) datasets. In the supplementary, we demonstrate that our model can work in a feed-forward fashion on a identity-specific retrieval dataset.
Finally, we discuss our findings and limitations in Sec.~\ref{subsec:discussion}. Qualitative results can be found in Figure~\ref{fig:retrieval}.

\subsection{Ablation study}
\label{subsec:ablations}
In this section, we evaluate our design choices on the evaluation section of the `this-is-my' dataset. As we want to obtain finer-grained insight into our model's performance, we compute what we call true R@5 (or tR@5, defined as a number of correct examples retrieved in top-5 over a number of all positives) and precision-at-5 (P@5; the proportion of positive examples retrieved in top-5).
Note that maximum theoretical tR@5=$40.3$. 

Table~\ref{tbl:ablations}~(a) shows that our model significantly outperforms naive CLIP baselines. Table~\ref{tbl:ablations}~(b) explores the effects of our modelling choices discussed in Sec.~\ref{sec:method} on the downstream performance. It is notable that caption augmentation plays a significant role in achieving good results ($+10.9$ precision points). While localisation plays only a minor role in the overall result, it allows us to achieve similar results with a lower number of frames (Tbl.~\ref{tbl:ablframes}).


\begin{table}[h]
    \centering
    \caption{Ablations on eval split of `this-is-my'~\cite{yeh2023meta} dataset.}
    \label{tbl:ablations}
    \begin{subtable}{0.4\textwidth}
        \centering
        \begin{tabular}{@{}lcc@{}}
        \toprule
        Method & \begin{tabular}[c]{@{}c@{}}tR@5\\ {\scriptsize(max 40.3)}\end{tabular} & P@5 \\ \midrule
        {\color[HTML]{656565} text -- generic} & {\color[HTML]{656565} 12.3} & {\color[HTML]{656565} 56.1} \\
        {\color[HTML]{656565} text -- specific} & {\color[HTML]{656565} 11.8} & {\color[HTML]{656565} 58.3} \\
        {\color[HTML]{656565} image} & {\color[HTML]{656565} 15.3} & {\color[HTML]{656565} 63.5} \\
        {\color[HTML]{656565} text + image} & {\color[HTML]{656565} 18.1} & {\color[HTML]{656565} 67.7} \\
        ours & 33.7 & 87.2 \\ \bottomrule
        \end{tabular}
        \vspace{2mm}
        \caption{Baseline results. Results in {\color[HTML]{656565} grey} denote CLIP~\cite{radford2021learning} baseline.}
    \end{subtable}
    \begin{subtable}{0.5\textwidth}
        \centering
        \begin{tabular}{@{}lcc@{}}
        \toprule
        Ablation & \begin{tabular}[c]{@{}c@{}}tR@5\\ {\scriptsize(max 40.3)}\end{tabular} & P@5 \\ \midrule
        Ours & 33.7 & 87.2 \\
        w/o Reg Loss & 29.1 & 79.1 \\
        w/o Caption Augmentation & 26.0 & 76.3 \\
        w/o Localisation & 31.9 & 83.5 \\
        w/o Pre-Training & 27.7 & 77.6 \\
        w/o Init-Scheme & 32.9 & 86.2 \\ \bottomrule
        \end{tabular}
        \vspace{2mm}
        \caption{Ablating various model components.}
    \end{subtable}
\end{table}

\begin{table}[t]
\centering
\caption{The performance of the method depends on a number of query images. Using local features reduces the amount of template images necessary. Results on the eval split of `this-is-my'~\cite{yeh2023meta} dataset.}
\label{tbl:ablframes}
\begin{tabular}{lcccc}
\toprule
\multirow{2}{*}{\begin{tabular}[c]{@{}l@{}}\#template\\ imgs\end{tabular}} & \multicolumn{2}{c}{with localisation} & \multicolumn{2}{c}{no localisation} \\ \cline{2-5} 
 & \multicolumn{1}{c}{\begin{tabular}[c]{@{}c@{}}tR@5\\{\scriptsize(max 40.3)}\end{tabular}} & \multicolumn{1}{c}{P@5} & \multicolumn{1}{c}{\begin{tabular}[c]{@{}c@{}}tR@5\\ {\scriptsize(max 40.3)}\end{tabular}} & \multicolumn{1}{c}{P@5} \\ \hline
1 & 31.9 & 84.5 & 28.6 & 77.8 \\
3 & 33.7 & 87.2 & 30.4 & 81.9 \\
5 & 33.6 & 87.2 & 31.9 & 83.5 \\
10 & 33.2 & 87.0 & 32.4 & 84.8 \\ \hline
\end{tabular}
\end{table}

\subsection{this-is-my}
\label{subsec:thisismyRes}
Results on this-is-my datasets are reported in Table~\ref{tbl:thisismy}. Our model using an image (randomly selected and held out from the training set) comfortably outperforms all other methods. It is notable that, although the text encoding is computed using an average of the template training images, using an image as a query as opposed to the text yields better results.

\begin{table}[t]
\centering
\caption{Results on the test set of `this-is-my'  dataset~\cite{yeh2023meta}. Included baseline use either text features or linear combination of image and text CLIP features for retrieval. `txt' denotes queries in plain text as seen in Fig.~\ref{fig:retrieval}. `*' denotes our reproduction of the baseline.}
\label{tbl:thisismy}
\begin{tabular}{@{}lcccc@{}}
\toprule
\multirow{2}{*}{Method} &  \multicolumn{2}{c}{Context} & \multicolumn{2}{c}{Generic} \\  
   & MRR & R@5 & mAP & MRR \\ \midrule
CLIP baseline~\cite{yeh2023meta} (txt) & 30.8 & 36.7 & 16.6 & 44.2 \\
CLIP baseline~\cite{yeh2023meta} (img+txt)  & 20.9 & 23.3 & 51.7 & 82.9 \\
CLIP*~\cite{radford2021learning} (img+txt) & 24.3 & 28.9 & 52.4 & 83,4 \\
Thisismy~\cite{yeh2023meta}  & 42 & 50.7 & 56.4 & 87.4 \\
Ours & \textbf{43.1} & \textbf{52.0} & \textbf{58.4} & \textbf{88.3} \\ 
\bottomrule
\end{tabular}
\end{table}


\subsection{DeepFashion2}
\label{subsec:deepfashionRes}

Our results on DeepFashion2~\cite{DeepFashion2} show marginal improvement over previous methods. DeepFashion2 is also the only dataset where caption augmentation did, in fact, cause adverse effects (54.7/78.2 with and 55.1/78.9 w/o). We hypothesise this is due to the relative simplicity of the object (e.g. `white skirt`) when compared to more complex descriptions of humans or particular objects in `this-is-my' dataset. Full results can be seen in Table~\ref{tbl:df2}.

\begin{table}
\centering
\caption{Results on `DeepFashion2' dataset~\cite{DeepFashion2}, personalization split as defined by~\cite{eccv2022_palavra_cohen}. Note that~\cite{eccv2022_palavra_cohen} use ViT-B/32 instead of ViT-B/16. Results in {\color[HTML]{656565} grey} denote CLIP~\cite{radford2021learning} baseline. `txt' denotes queries in plain text as seen in Fig.~\ref{fig:retrieval}.}
\label{tbl:df2}
\begin{tabular}{@{}lcccc@{}}
\toprule
 & \multicolumn{2}{c}{Context} & \multicolumn{2}{c}{Generic} \\
\multirow{-2}{*}{Method} & MRR & R@5 & mAP & MRR \\ \midrule
{\color[HTML]{656565} txt} & {\color[HTML]{656565} 21.2} & {\color[HTML]{656565} 23.4} & {\color[HTML]{656565} 9.0} & {\color[HTML]{656565} 17.5} \\
{\color[HTML]{656565} img} & {\color[HTML]{656565} 14.5} & {\color[HTML]{656565} 17.6} & {\color[HTML]{656565} 20.9} & {\color[HTML]{656565} 43.9} \\
{\color[HTML]{656565} img + txt} & {\color[HTML]{656565} 21.0} & {\color[HTML]{656565} 26.9} & {\color[HTML]{656565} 21.7} & {\color[HTML]{656565} 43.6} \\
PALVARA~\cite{eccv2022_palavra_cohen} & 28.4 & 39.2 & - & - \\
this-is-my~\cite{yeh2023meta} & 38.4 & 51.4 & 53.4 & 77.7 \\
Ours & 38.5 & 51.8 & 54.7 & 78.2 \\ \bottomrule
\end{tabular}
\end{table}

\subsection{Discussion and Limitations}
\label{subsec:discussion}
We demonstrate that our model learns image-to-text mapping with less examples and achieving higher performance than all other personalization methods. 
One limitation shared with most previous work is that the model has to be fine-tuned to learn the token for each instance (though this only has to be done once). While our method could in theory address this issue, we reserve these experiments for future work. Furthermore, although our model is much class agnostic in comparison to previous work, there is still bias and lack of cross-domain robustness due underlying use of CLIP~\cite{Shtedritski23}.



\section{Conclusion}
We present a conceptually simple and effective method for learning personalised tokens in VLMs using image-to-text mapping called $\pi$-map. It is highly capable in personalised text-to-image, text-to-video, and image-to-image retrieval, outperforming all prior work on three personalization benchmarks while requiring only a few examples to fully personalise the embedding. In future work, we hope to expand on our method and develop a new, larger benchmark for personalised retrieval. 

%% file: supp_sec/1_localfeatures.tex
\section{Importance of Local Features}
\begin{figure}[h]
    \centering
    \begin{subfigure}[t]{0.47\textwidth}
        \centering
        \includegraphics[width=0.9\textwidth]{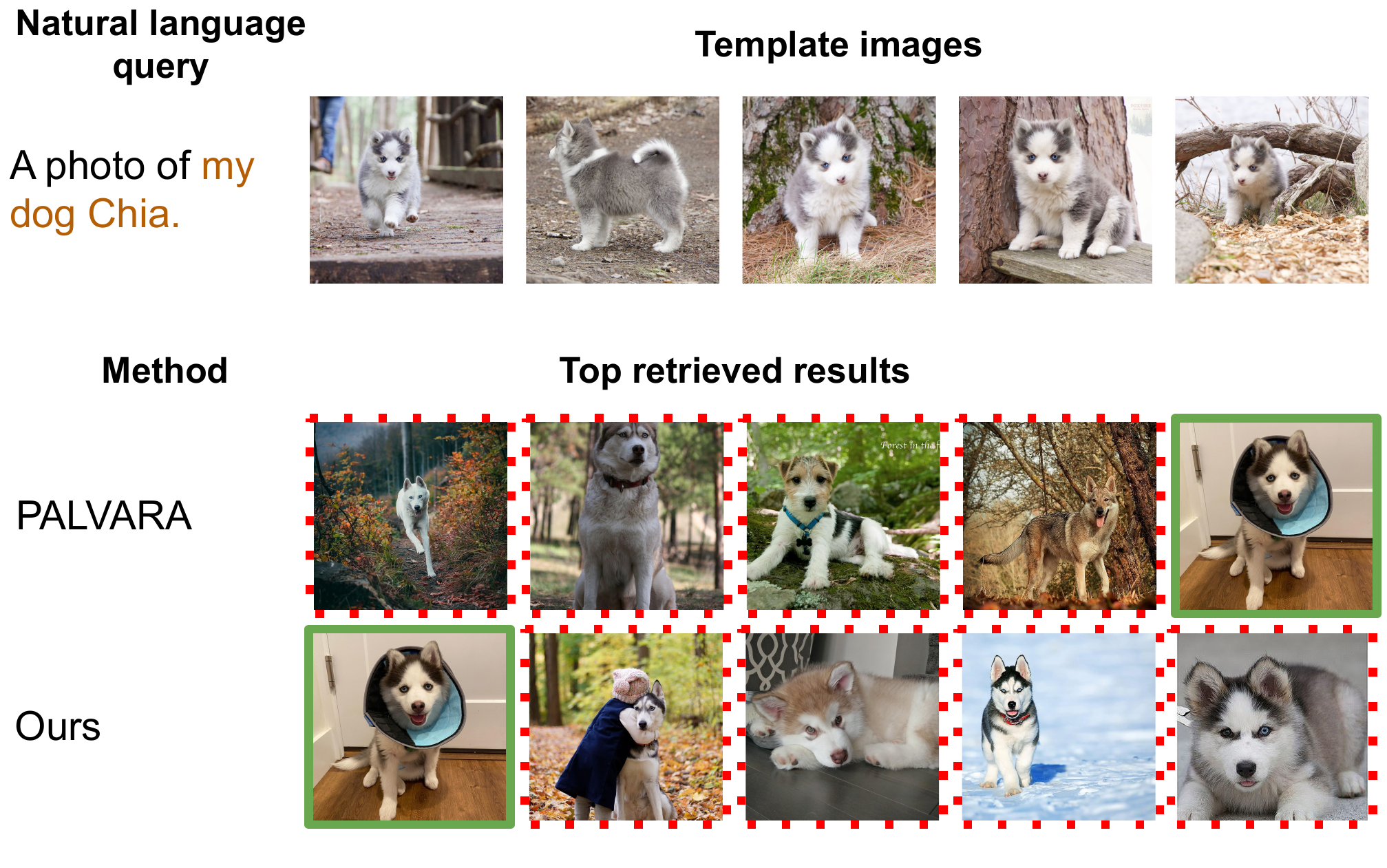}
        \caption{}
    \end{subfigure}
    \begin{subfigure}[t]{0.47\textwidth}
        \centering
        \includegraphics[width=0.9\textwidth]{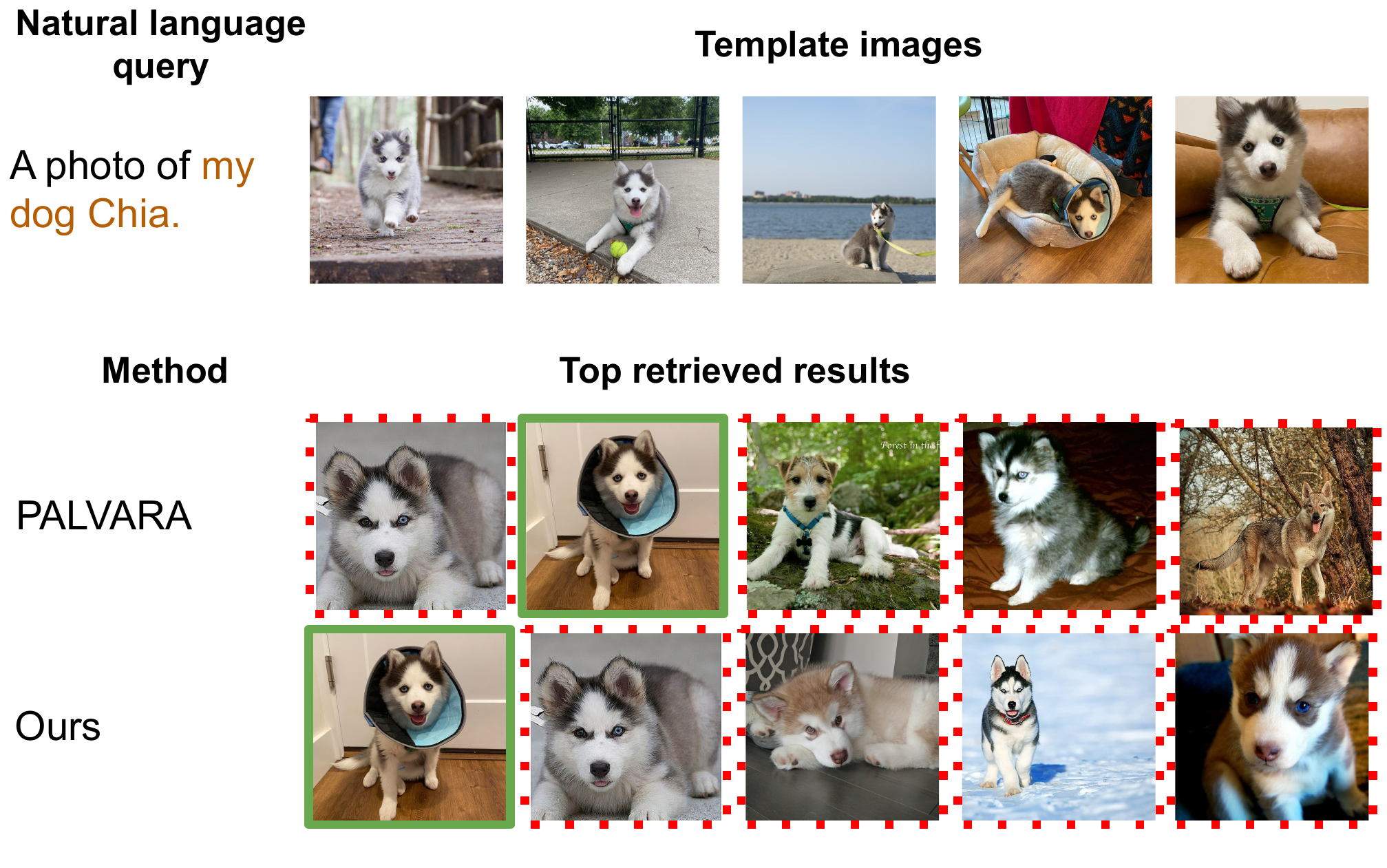}
        \caption{}
    \end{subfigure}
    \caption{Importance of using localised features: learning personalised features for `My dog Chia' from two different sets of template images. In \textit{a}) all template images come from the same time and place, while in \textit{b)}, the images are varied. Our method ranks the correct image first on both occasions, while PALVARA~\cite{eccv2022_palavra_cohen}  remains sensitive to the diversity of the template images.} 
    \label{fig:localisation}
\end{figure}

We investigate the quantitative importance of using local features for learning personalised embeddings in Tbl.~1 and Tbl.~2 of the main paper. To demonstrate the importance qualitatively, we learn personalised embedding with 5 different images of the dog `Chia' using our method and those of PALVARA~\cite{eccv2022_palavra_cohen}. We obtain 40 images with Google image search to use as hard negatives (prompts used: `a dog in a forest' and an `a small husky in a forest'), and display top-5 in Figure~\ref{fig:localisation}. Our method shows resilience to the type of background features, while PALVARA~\cite{eccv2022_palavra_cohen} seems to exploit additional biases (such as background) in the images.

%% file: supp_sec/2_results.tex
\section{Invariance to CLIP Models}
In the main paper, we present results only using OpenAI's CLIP (VIT-B/16)~\cite{radford2021learning} in order to compare fairly with state-of-the-art methods. To demonstrate that our method can work on various CLIP variants pre-trained on different datasets in a `plug-and-play' fashion, we use fixed setup described in the main body using personalised embeddings without image queries, and experiment with various CLIP variants provided by~\cite{openclip}. Our results (Table~\ref{tbl:clips}) demonstrate that our method can be applied in a plug-and-play fashion. Most variants deviate only slightly from the baseline model (within $1.1$ recall point on `this-is-my' and `CiA' datasets). 

\begin{table}[]
\centering
\caption{Exploration of performance using various CLIP variants. All results are computed using text-only queries on the test sets of this-is-my~\cite{yeh2023meta}, CiA~\cite{Korbar22}, and DeepFashion2~\cite{DeepFashion2} datasets.}
\label{tbl:clips}

\begin{tabular}{lcccccc}
\hline
 & \multicolumn{2}{c}{\begin{tabular}[c]{@{}c@{}}\textbf{ThisIsMy}\\ \textbf{Context}\end{tabular}} & \multicolumn{2}{c}{\textbf{CiA}} & \multicolumn{2}{c}{\begin{tabular}[c]{@{}c@{}}\textbf{DeepFashion2}\\ \textbf{Context}\end{tabular}} \\
\multirow{-2}{*}{CLIP Variant} & MRR & R@5 & R@1 & R@5 & MRR & R@5 \\ \hline
VIT-B/16~\cite{radford2021learning} & 42.1 & 50.9 & 64.9 & 81.2 & 38.3 & 51.2 \\
{\color[HTML]{656565} ViT-B/32}~\cite{radford2021learning} & {\color[HTML]{656565} 41.6} & {\color[HTML]{656565} 50.4} & {\color[HTML]{656565} 64.7} & {\color[HTML]{656565} 81.1} & {\color[HTML]{656565} 38.3} & {\color[HTML]{656565} 51.0} \\
{\color[HTML]{656565} ViT-L/14}~\cite{openclip} & {\color[HTML]{656565} 42.4} & {\color[HTML]{656565} 51.0} & {\color[HTML]{656565} 65.4} & {\color[HTML]{656565} 81.4} & {\color[HTML]{656565} 38.6} & {\color[HTML]{656565} 51.5} \\
{\color[HTML]{656565} ViT-H/14}~\cite{openclip} & {\color[HTML]{656565} 42.7} & {\color[HTML]{656565} 51.5} & {\color[HTML]{656565} 65.5} & {\color[HTML]{656565} 81.4} & {\color[HTML]{656565} 38.9} & {\color[HTML]{656565} 52.0} \\
{\color[HTML]{656565} \begin{tabular}[c]{@{}l@{}}ViT-SO400M/14\\ (siglip)~\cite{openclip, zhai2023sigmoidlosslanguageimage}\end{tabular}} & {\color[HTML]{656565} 43.4} & {\color[HTML]{656565} 52.0} & {\color[HTML]{656565} 66.0} & {\color[HTML]{656565} 82.3} & {\color[HTML]{656565} 39.4} & {\color[HTML]{656565} 53.7} \\ \hline
\end{tabular}
\end{table}



%% file: supp_sec/4_datasets.tex
\section{Dataset examples}
\begin{figure}
    \centering
    \begin{subfigure}[t]{.33\textwidth}
    \centering
        \includegraphics[width=\textwidth]{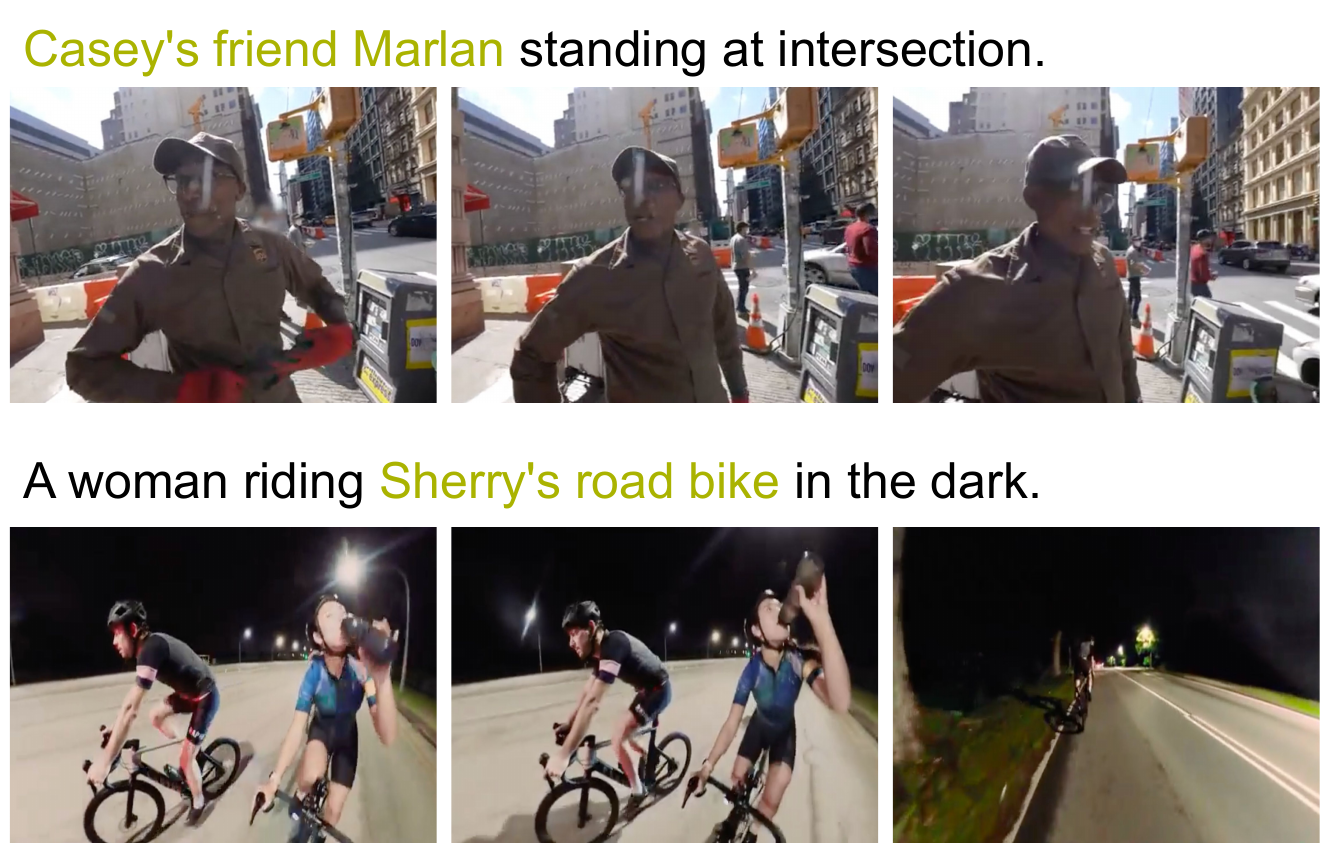}
        \caption{`this-is-my' dataset~\cite{yeh2023meta}.}
    \end{subfigure}
    \begin{subfigure}[t]{.3\textwidth}
    \centering
        \includegraphics[width=.73\textwidth]{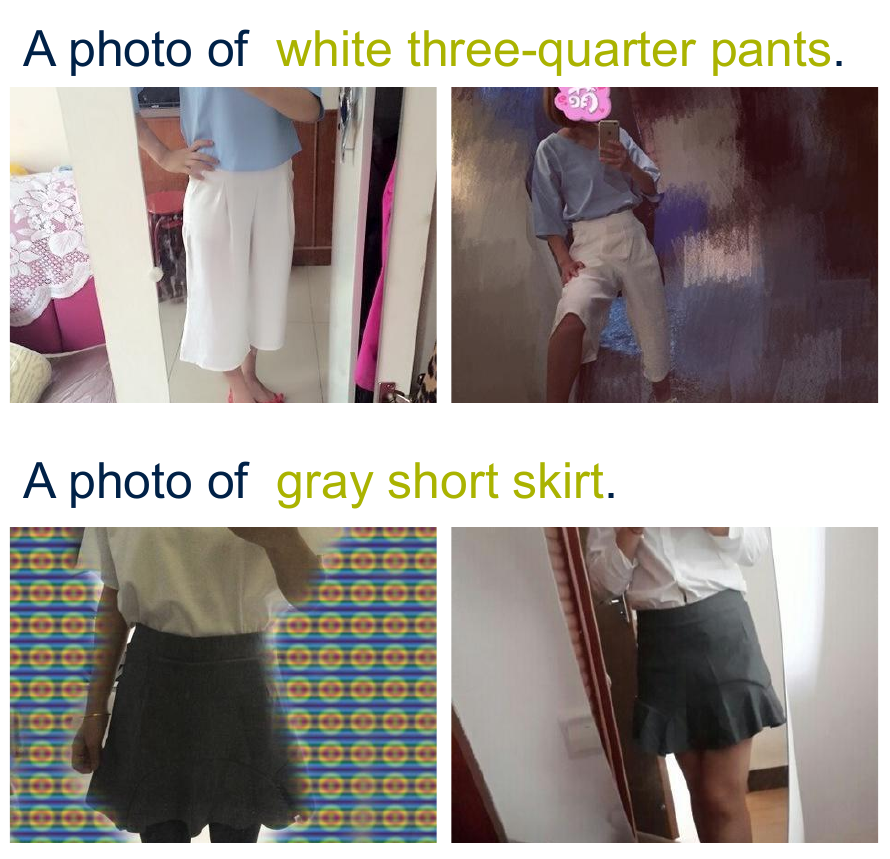}
        \caption{DeepFashion2 dataset~\cite{DeepFashion2}.}
    \end{subfigure}
    \caption{Examples from our evaluation datasets.}
    \label{fig:datasets}
    \vspace{-5pt}
\end{figure}

Samples from the evaluation datasets are given in Fig.~\ref{fig:datasets}.

